\documentclass[runningheads]{llncs}
\usepackage[T1]{fontenc}
\usepackage{graphicx,verbatim}
\usepackage{adjustbox}
\usepackage{xcolor}
\usepackage{enumerate}
\usepackage{notoccite}
\usepackage{multirow}
\usepackage{amsmath}
\usepackage{amssymb}
\usepackage{subfig}
\usepackage{tabularx}
\usepackage{makecell}
\usepackage{hyperref}
\usepackage{tikz, graphics, float, pgfplots}
\usepackage{graphicx}
\usepackage{pifont}
\usepackage{graphicx,verbatim}
\usepackage{siunitx}
\usepackage{orcidlink}
\RequirePackage{xspace}
\makeatletter
\DeclareRobustCommand\onedot{\futurelet\@let@token\@onedot}
\def\@onedot{\ifx\@let@token.\else.\null\fi\xspace}

\def\etal{\emph{et al}\onedot}
\makeatother
\def\method{DeepAf}
\usepackage{cuted} 
\newcommand{\xpm}[1]{{\tiny$\pm#1$}}
\usepackage{tikz}

\begin{document}
\title{\method{}: One-Shot Spatiospectral Auto-Focus Model for Digital Pathology}

\author{Yousef Yeganeh\orcidlink{0000-0002-0000-0000} \inst{1,2} \and Maximilian Frantzen\inst{1} \and Michael Lee\inst{3} \and \\ Kun-Hsing Yu\orcidlink{0000-0001-9892-8218}\inst{4,5,6,7} \and Nassir Navab\orcidlink{0000-0002-6032-5611} \inst{1,2} \and Azade Farshad\orcidlink{0000-0002-1080-1587} \inst{1,2}}

\authorrunning{Y. Yeganeh et al.}
\institute{
    Chair for Computer Aided Medical Procedures (CAMP), TU Munich, Germany \\
    \and Munich Center for Machine Learning (MCML) \\
    \and Southern Taiwan University of Science and Technology, Taiwan \\
    \and Dep. of Biomedical Informatics \and Pathology, Brigham and Women's Hospital, Boston, MA, USA \\
    \and Harvard Data Science Initiative \and Kempner Institute for the Study of Natural and Artificial Intelligence, Harvard University, Cambridge, MA, USA \\
    \email{y.yeganeh@tum.de}
}
\maketitle
\begin{abstract}
While Whole Slide Imaging (WSI) scanners remain the gold standard for digitizing pathology samples, their high cost limits accessibility in many healthcare settings. Other low-cost solutions also face critical limitations: automated microscopes struggle with consistent focus across varying tissue morphology, traditional auto-focus methods require time-consuming focal stacks, and existing deep-learning approaches either need multiple input images or lack generalization capability across tissue types and staining protocols. We introduce a novel automated microscopic system powered by \method{}, a novel auto-focus framework that uniquely combines spatial and spectral features through a hybrid architecture for single-shot focus prediction.
The proposed network automatically regresses the distance to the optimal focal point using the extracted spatiospectral features and adjusts the control parameters for optimal image outcomes. Our system transforms conventional microscopes into efficient slide scanners, reducing focusing time by 80\% compared to stack-based methods while achieving focus accuracy of $0.18$ $\mu m$ on same-lab samples—matching the performance of dual-image methods ($0.19$ $\mu m$) with half the input requirements. \method{} demonstrates robust cross-lab generalization with only 0.72\% false focus predictions and 90\% of predictions within the depth of field. Through an extensive clinical study of 536 brain tissue samples, our system achieves 0.90 AUC in cancer classification at $4\times$ magnification, a significant achievement at lower magnification than typical 20$\times$ WSI scans. This 
results in a comprehensive hardware-software design enabling accessible, real-time digital pathology in resource-constrained settings while maintaining diagnostic accuracy.
\end{abstract}
\section{Introduction}\label{sec:intro}
Through the digitization of tissue samples in pathology, machine learning models trained on these images have transformed our ability to detect and classify diseases \cite{digitize_breast}. However, the fundamental challenge of capturing high-quality microscopic images at speed remains unsolved. Although tissue samples are sliced into micrometer-thin sections, due to the fine focal length of microscopic lenses, they retain complex 3D morphological structures that create continuously varying optimal focal planes during the scanning process \cite{Li2020.04.11.037473}. Capturing images in the optimal focal plane is critical to achieving high-quality and detailed images, and neglecting this leads to image quality degradation that can severely impact diagnostic accuracy and increase healthcare costs through repeated scans or potential misdiagnoses \cite{article23}.

Auto-focus methods can be broadly classified into two categories: traditional and learning-based. 
Contrast-based traditional methods assess focus through gradient-based algorithms \cite{yeo_tenenbaum_1993,santos_evaluation_1997,brenner_automated_1976,subbarao_energy_labplace_1993} —all requiring time-intensive capture of complete focal stacks. Later, optimization-based approaches \cite{yazdanfar_simple_2008} approximated the Brenner gradient with Lorentzian functions using sparse focal positions, while others \cite{nayar_shape_1994,wang_compact_2015} employed Gaussian models. Yet these methods falter in noisy environments due to local maxima issues. Phase-based methods leveraged dual-pixel sensors for disparity-based depth computation \cite{herrmann_learning_2020}, but struggled with the fundamental complexity of depth-disparity modeling \cite{garg_disparity}. Deep learning approaches have transformed auto-focusing in pathology by leveraging CNNs. Early methods focused on single-domain feature extraction. Wei et al. \cite{wei_neural_2018} pioneered CNN-based focus prediction for time-lapse cell microscopy, while Jiang et al. \cite{jiang_transform-_2018} introduced autocorrelation in their residual architecture. Recent approaches \cite{dastidar_whole_2020,liao_deep_2021} have adapted CNN architectures from computer vision, such as MobileNet.
These learning-based methods \cite{gan} significantly outperform traditional approaches in speed and noise robustness, though they often struggle to generalize across tissue types and staining protocols. Recent developments on improving generalization propose a sample-invariant CNN scoring function \cite{deep-focus}, Kernel Distillation using paired samples for training but single-shot inference \cite{yun}, or tackling focused image reconstruction under incoherent lighting conditions \cite{Ding}.

Automating conventional microscopes for pathology remains a significant challenge, with most existing work focusing on narrow, specialized applications. Chow et al. \cite{chow_automated_2006} enabled multi-region mosaic imaging for multi-photon microscopes, while \cite{aoyama2021view} developed automation for micro-injection procedures. Li et al. \cite{li_lowcost_2019} implemented a three-motor system for parasite detection, but relied on proprietary software, limiting reproducibility and customization. Collins et al. \cite{Collins:20} contributed to accessibility through 3D-printed frameworks for motor integration, yet left the crucial auto-focus challenge unaddressed.

\begin{figure}[tb]
\centering
\includegraphics[width=\linewidth]{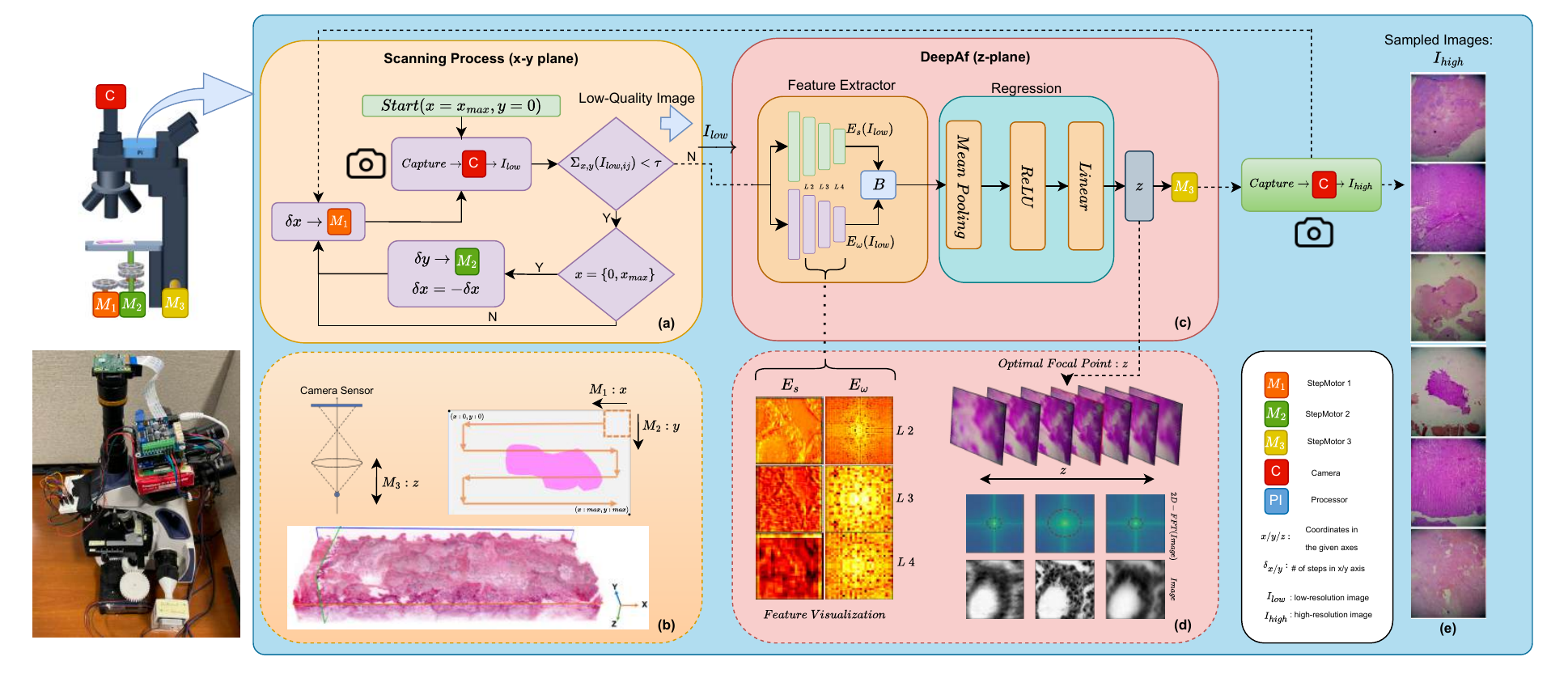}
\caption{\textbf{Left:} Implemented prototype and the schematic of the microscopic setup. Step motors $M_1$, $M_2$, $M_3$ control slide movement using processor $PI$ along $x,y,z$ respectively (a), with $M_3$ controlling the focus position (b). At each step, a low-resolution image $I_{low}$ is captured by the camera ($C$), and the non-empty images are fed to the \textit{\method{}} network to adjust the focus (c). Finally, the high resolution images $I_{high}$ are obtained with the correct focus (d).}
\label{fig:full_pipeline}
\end{figure}

This work proposes \method{} (\textbf{Deep} \textbf{A}uto\textbf{f}ocus)\footnote{\textbf{Project Page}: \url{https://deepautofocus.github.io/}}, a novel deep learning-based auto-focus framework based on the learnable features from Spatial and Spectral encoders in a regression model to predict the optimal focal distance without additional stacking or multiple views. We show that a single out-of-focus image contains enough information to infer the optimal focus due to the relationship between defocus and frequency domain characteristics. Defocusing creates unique signatures in the cut-off frequency and spectral distribution, which correlate with the defocus distance. Based on the observations that the cut-off frequency of a Fourier-transformed image indicates its distance to the optimal focal plane \cite{jiang_transform-_2018}, we developed an automated microscopy system with a hybrid architecture inspired by Y-Net \cite{farshad_y-net_2022}. Our model simultaneously learns features from both spatial and frequency domains, enabling better generalization across different tissue types. This approach requires only a single input image to predict the optimal focus point, significantly reducing focusing time compared to traditional methods.
We developed an efficient motorized microscopic design based on a conventional manual microscope \cite{micSwfit}. \autoref{fig:full_pipeline} depicts the auto-scanning setup of the microscope. Our framework incorporates \method{} to efficiently capture high-quality images from histopathology slides. For clinical validation, we used the framework to create a dataset of brain tissue samples and demonstrated its effectiveness through automated cancer classification. This validation shows the potential of our approach for real-world clinical applications.
In summary, our key contributions are: (1) \method{}: A novel single-shot auto-focus approach effectively utilizing features from the spatial and spectral encoder in a regression model, demonstrating superior generalization across diverse tissue types and staining protocols in a single-shot without additional views or resampling, (2) An open-source, automated microscopy system that transforms conventional microscopes into efficient slide scanners, maintaining optimal focus throughout the imaging process, (3) A comprehensive dataset of brain tissue samples captured using our system, showcasing its ability to consistently produce high-quality digital pathology images, (4) Experimental validation through automated cancer classification, demonstrating the system's practical utility in clinical diagnostics.
\section{Method}\label{sec:method}
\subsection{Microscopic System}
The digitization pipeline consists of four sequential stages:
(1) Slide Scanning: Following DICOM standards, scanning begins from the bottom right corner and follows a predefined trajectory (\autoref{fig:full_pipeline}-b). The system acquires low-resolution (1280×720) images for initial tissue detection, optimizing computational efficiency. (2)  Tissue Detection: Simple thresholding to skip empty spaces is widely used in histopathology image analysis, such as Otsu thresholding \cite{bandi2018detection}. Similarly, we apply thresholding in the HSV color space to filter empty regions:
$$\frac{1}{N} \sum_{i=1}^{N} V_i > \tau$$
where $V_i$ represents pixel values in the Value channel of HSV and $\tau$ is the threshold value. (3) Auto-focus: Our spatiospectral network performs single-shot focus prediction, efficiently determining the optimal focal plane for the detected tissue region. The model's compact size enables real-time inference on the Raspberry Pi CPU. (4) High-Resolution Capture: The system captures the final image at 4056×3040 resolution at the predicted focal position $\hat{z}$. The scanning step size in the $x$-$y$ plane critically impacts system performance. Small steps increase overlap and scanning time, while large steps risk missing tissue regions. Optimal step size selection balances coverage completeness with scanning efficiency.

Our automated microscopy system is built around a SWIFT-380t microscope \cite{micSwfit}. The system consists of three key components: (1) A motorized stage with three stepper motors: M1 and M2 control x-y slide positioning, and M3 adjusts focus in the z-direction with 0.002mm precision. Custom 3D-printed gears translate motor motion to the microscope bed. (2) A dual-mode camera mounted directly above the objective lens eliminates parallax correction requirements. The camera operates at either low-resolution ($I_{low}$) for rapid tissue detection or high-resolution ($I_{high}$) for final image capture. (3) A Raspberry Pi control system with dual motor control HAT (Hardware Attached Top) coordinates motor movements and image acquisition based on auto-focus predictions.

\subsection{\method{}}
Let $\mathcal{M} = (X,Y,Z)$ define the microscope coordinate system where $Z$ represents the focus direction. For an input image $I \in \mathbb{R}^{H \times W \times C}$, the system comprises:
A focus prediction function $f: \mathbb{R}^{H \times W \times C} \rightarrow \mathbb{R}$ defined as:
$f(I) = R(B(E_s(I), E_\omega(I)))$
, where $E_s, E_\omega$ are spatial and spectral encoders respectively, $B$ is a bottleneck layer, and $R$ is the regression head.
A motorized control system with precision $\delta_z$mm adjusts the focal plane according to $f(I)$. The system captures images at position $(x,y) \in X \times Y$ with focus position $\hat{z} = f(I)$ to maintain optimal focus throughout scanning (\textit{cf.} \autoref{fig:full_pipeline}-a). The dual encoder architecture (\autoref{fig:full_pipeline}-c) consists of: 1) Spatial Encoder: $E_s: \mathbb{R}^{H \times W \times C} \rightarrow \mathbb{R}^{h \times w \times d_s}$ following U-Net architecture \cite{ronneberger_u-net_2015} extracting spatial features via four hierarchical convolutional layers. 2) Spectral Encoder: $E_\omega: \mathbb{R}^{H \times W \times C} \rightarrow \mathbb{R}^{h \times w \times d_\omega}$ comprising four sequential FFC blocks \cite{fft-blocks} that extract frequency domain features. The concatenated features $[E_s(I), E_\omega(I)] \in \mathbb{R}^{h \times w \times (d_s + d_\omega)}$ are processed by bottleneck $B$, followed by a regression head $R$ consisting of 2D average pooling and a linear layer to predict the optimal focus position $\hat{z}$ along the $Z$ axis of the microscope coordinate system.  Unlike previous approaches such as \cite{jiang_transform-_2018}, our architecture learns to extract both spatial and spectral features directly from the input image, eliminating preprocessing overhead. The modular design enables independent evaluation of spatial and spectral contributions to focus prediction. We train the network using smooth $\mathcal{L}1$ loss to handle the extensive range of focal values while maintaining gradient stability \cite{smoothL1loss}.

As demonstrated in \autoref{fig:full_pipeline}-d, in the 2D FFT image, the power spectrum analysis reveals distinct patterns between in-focus and out-of-focus images. Out-of-focus images significantly attenuate low-frequency components near the spectrum's center, while in-focus images display enhanced low-frequency amplification and higher cut-off frequencies, as also observed in \cite{jiang_transform-_2018}. This consistent relationship between focus quality and frequency distribution suggests that spectral features could provide robust indicators for auto-focus systems, potentially offering better generalization across different tissue types and imaging conditions.

\section{Experiments}
\noindent \textbf{Datasets}\label{ss:results-auto-focus-datasets}
To evaluate our method, we employ the Incoherent dataset by Jiang \etal \cite{jiang_transform-_2018}, which comprises two distinct test sets. The first test set contains tissue samples prepared by the same lab as the training data, while the second one includes specimens from a different lab. The two test sets exhibit significant differences in their color distribution, which serves as a good indicator of the generalization capabilities of the models. All images were captured at magnification level 20$\times$ with a depth of field (DoF) of 1$\mu$. For both datasets, during training, all images are divided into tiles of size 224$\times$224 pixels. 
Furthermore, for the case study of our microscopic system, we train the auto-focus network on 406 brain tissue samples from the Brigham and Women’s Hospital in Boston and the University of Pennsylvania. For each sample, we create a focal stack of 1000 slices such that 500 images are below and above the optimal focal plane. Here, we capture the tissue at magnification level 4$\times$ with a DoF of 60$\mu$ to ensure faster scanning times in the subsequent case study.

\vspace{5pt} \noindent \textbf{Training and Evaluation}\label{ss:results-auto-focus-train-eval}
During training, we optimize the model on each individual patch with a total of 130K patches. For hyperparameter optimization, we further split the data into $80\%$ training and $20\%$ validation. The test data consists of 700 patches. During testing, the median of all patches of one image serves as the final prediction for evaluation.
We train all models with a batch size of 32, a learning rate of 8e-4, a weight decay of 0.006, 100 epochs, and the Adam optimizer. Moreover, all models presented in this work have been trained with data augmentation, namely, channel-wise normalization, random erasing, Gaussian blur, random perspective, random auto contrast, and color jittering. We report the model performance as focus error (FE) computed by the mean absolute error between the predicted and optimal focal distance and its standard deviation.

\begin{table}[tb]
\centering
\caption{Comparison of SOTA auto-focus methods to ours on incoherent dataset \cite{jiang_transform-_2018} test set.}
\begin{tabular}{ccccc}
\hline
\multirow{2}{*}{Method} & \multirow{2}{*}{Params.} & \# of input  & \multicolumn{2}{c}{FE $\downarrow$} \\ \cline{4-5}
& & images &  Same protocol  &  Diff. protocol \\
\hline
\thead{Dastidar \etal \cite{dastidar_whole_2020}} & 3.5M  & 2 & $0.19$\xpm{0.18} & $\textbf{0.25}$\xpm{\textbf{0.26}}\\ 
\thead{Jiang \etal \cite{jiang_transform-_2018}}  & 10.8M & 1 & $0.46$\xpm{0.34}         & $0.53$\xpm{0.59}\\
\thead{Chen \etal \cite{chen2024microscope}} & 4.2M & 1 & $0.21$\xpm{0.21} & $0.44$\xpm{0.50}\\
\hline
\method{} Spatial (Ours) & 4.7M & 1 & $\textbf{0.18}$\xpm{\textbf{0.17}} & $0.39$\xpm{0.50} \\
\method{} Spatiospectral (Ours) & 4.2M &  1 & $\textbf{0.18}$\xpm{\textbf{0.17}} & \underline{$0.32$\xpm{0.36}}\\ 
\end{tabular}
\label{tab:comparison-open-source-data-state-of-the-art}
\end{table}

\subsection{Autofocus Results}\label{ss:results-auto-focus-results}
\noindent \textbf{Comparison to SOTA} 
\autoref{tab:comparison-open-source-data-state-of-the-art} shows our model's performance compared to previous auto-focus methods. All reported results from previous work are taken from the original publications. On the same protocol data, our spatiospectral model achieves the best overall focus error while using significantly fewer parameters compared to \cite{jiang_transform-_2018} and taking only one input image instead of two as in \cite{dastidar_whole_2020}. On the different protocol data, the spatiospectral network performs significantly better than \cite{jiang_transform-_2018} and only exhibits a slightly bigger focus error than reported by \cite{dastidar_whole_2020}, outperforming the single-shot SOTA auto-focus model \cite{chen2024microscope} by Chen \etal. It is noteworthy to mention that our method only takes one input image while \cite{dastidar_whole_2020} takes two input images to predict the distance to the optimal focal plane. The choice of only taking one input image is motivated by higher inference times, which is critical to allow for efficient scanning of histopathology slides in a clinical setting. Our method shows promising generalization capabilities compared to previous single-shot auto-focus methods \cite{jiang_transform-_2018,chen2024microscope}.

\begin{figure}[tb]
\centering
\begin{tabular}{cc}
    \includegraphics[width=0.48\linewidth]{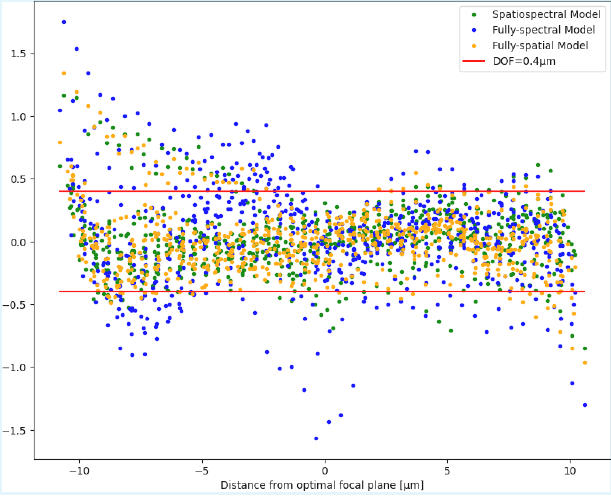}  &  \includegraphics[width=0.47\linewidth]{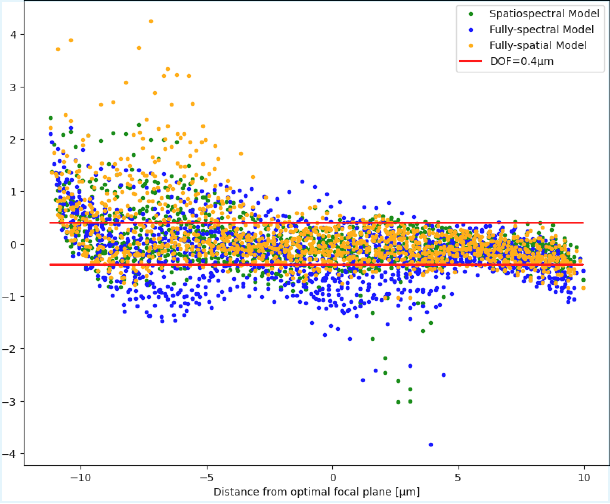}
\end{tabular}
\caption{\textbf{Focal Distance from Optimal Focal Point.} Left: data from the same protocol, Right: data from the different protocol.}
\label{fig:spatiospectral-same-predictions}
\end{figure}

\vspace{5pt} \noindent \textbf{Ablation Study} \autoref{tab:comparison-open-source-data-own-models} shows the effect of different components in our network on the open dataset (20$\times$ magnification) \cite{jiang_transform-_2018} and our own data (4$\times$ magnification). Here, FD, DoF, and FE denote the predictions with false directions, depth of field, and focus error, respectively. For the public dataset, we can see that the spatiospectral and fully-spatial models achieve the same performance on the same protocol data. However, the spatiospectral network outperforms the fully-spatial one on the different protocol data, exhibiting a better generalization in different lab environments. Moreover, the fully-spectral encoder shows the highest error rate for both datasets. On the one hand, these observations indicate that including features from the spectral domain can enhance the generalization performance of the auto-focus. On the other hand, the network seems to learn essential features from the spatial domain, as evidenced by the good performance of the fully-spatial model for the same protocol data and the overall bad performance of the fully-spectral model.
In the case of our data, the fully spatial model slightly outperforms the spatiospectral network. Since we only have data from one lab for this dataset, we cannot check for the generalization capabilities of the individual models. We also see that the spatiospectral model only predicts 0.72\% of the cases in the wrong focus direction (above or below the optimal focal plane), indicating that it can solve the focus ambiguity problem with just one input image. Moreover, approximately 90\% of all predictions lie inside the DoF. From a practical point of view, this implies that at least 90\% of the captured images based on this auto-focus model appear visibly sharp to the human eye. We also show a visualization of the focal error distribution given different distances from the optimal focal point using the data from the same and different protocols in \autoref{fig:spatiospectral-same-predictions} for different models. As can be seen, while the spatiospectral model generally handles the different distances from the same protocol data well, the fully spatial model can generalize better to data with a different protocol. We assume that this can be due to the larger changes in the frequency domain between different data distributions.

\begin{table}[tb]
\centering
\caption{\textbf{Ablation Study.} Focus Error comparison between different encoders on the Incoherent dataset (same and different protocols) \cite{jiang_transform-_2018} and our curated dataset. Spatial: Two spatial encoders, Spectral: Two spectral encoders, Spatiospectral: One spatial and one spectral encoder.}
\resizebox{\linewidth}{!}{
\begin{tabular}{ccccccccc}
\hline
\multirow{4}{*}{Network} & \multirow{4}{*}{Params.} & \multicolumn{2}{c}{FD $\downarrow$}  & \multicolumn{2}{c}{DoF $\uparrow$} & \multicolumn{3}{c}{FE $\downarrow$} \\\cline{3-3} \cline{4-6} \cline{7-9}
& & \thead{Same \\ 20$\times$ Magn.}  &  \thead{Diff. \\ 20$\times$ Magn.} & \thead{Same \\ 20$\times$ Magn.}  &  \thead{Diff. \\ 20$\times$ Magn.} & \thead{Same \\ 20$\times$ Magn.}   &  \thead{Diff. \\ 20$\times$ Magn.}  & \thead{Ours \\ 4$\times$ Magn.} \\
\hline
Spatial & 4.7M & 0.86\% & \textbf{1.22\%} & 89.24\% & 71.34\% & \textbf{0.18}\xpm{\textbf{0.17}} & 0.39\xpm{0.50}   & \textbf{5.40}\xpm{\textbf{6.06}}\\
Spectral & 3.6M & 1.29\% & 3.12\% & 73.03\% & 53.81\% & 0.29\xpm{0.26}  & 0.46\xpm{0.39} & 9.24\xpm{9.33} \\
Spatiospectral & 4.2M & \textbf{0.72\%}  & 1.60\% & \textbf{89.81\%}  & \textbf{73.78\%} & \textbf{0.18}\xpm{\textbf{0.17}} & \textbf{0.32}\xpm{\textbf{0.36}} & 6.67\xpm{7.51} \\
\hline
\end{tabular}
}
\label{tab:comparison-open-source-data-own-models}
\end{table}

\begin{figure}[tb]
\centering
\includegraphics[width=0.4\linewidth]{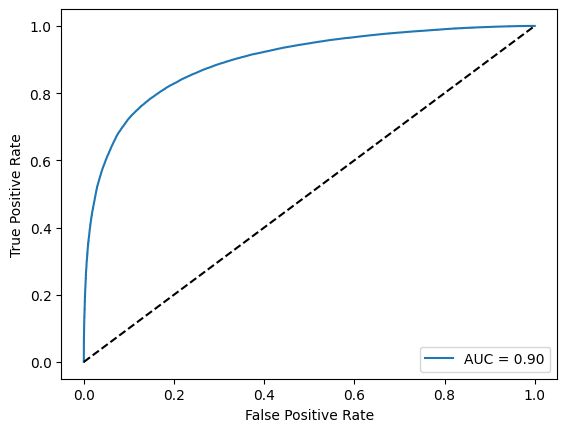}
\caption{\textbf{Classification Performance.} Receiver operating curve for the binary brain cancer classifier with an AUC score of 0.9.}
\label{fig:auroc-classifier}
\end{figure}

\subsection{Case study: Brain Tissue Slides Scanning and Classification}\label{ss:results-mic}
\noindent \textbf{Brain Tissue Dataset}
With the proposed scanning strategy and utilizing the spatiospectral auto-focus model, we scanned 536 histopathology slides containing brain tissue from Brigham and Women’s Hospital in Boston. The tissue samples comprise four different cancer subtypes, namely high-grade glioma, low-grade glioma, inflammatory, and normal tissue. As proof of concept, we used the 4$\times$ magnification objective lens of the microscope to have a bigger FoV and, thus, faster scanning times. The average scanning time for each slide is approximately 400 seconds, depending on the amount of tissue present. \autoref{fig:full_pipeline}-e illustrates some qualitative results of the acquired images.

\vspace{5pt} \noindent \textbf{Brain Cancer Classification}
We use the captured images to train a binary classification model. We define the high and low-grade glioma classes as the "cancer" class, while the inflammatory and normal samples are defined as the "normal" class. As seen in \autoref{fig:full_pipeline}-e, due to the high magnification rate, the images exhibit large radial distortions at the edges. Thus, each image is center-cropped to a size of 2000$\times$3000 pixels. During training, each tile and its corresponding label are considered individually, while at inference, the global mean of all tiles belonging to one tissue sample is computed as the final prediction.

\vspace{5pt} \noindent \textbf{Results}
The binary classifier achieves an AUC score of 0.90 and an F1 score of 0.83 in the 5-fold cross-validation setting. \autoref{fig:auroc-classifier} shows the corresponding ROC curve. It is noteworthy to mention that these results were achieved with just a magnification level of 4$\times$. We do not compare this result with previous works since they rely on WSIs, which are scanned at a magnification of 20$\times$, capturing much more detail of the tissue.
Nevertheless, this result indicates the high quality of the generated images of our microscopic system and their relevance and validity for automated cancer diagnosis in a clinical setting. 
\section{Conclusion}
In this work, we developed an automated robotic microscopic system for scanning histopathology glass slides while maintaining the optimal focus position during the process. A key component of this system is our deep auto-focus model, which shows superior generalization performance by only taking one input image. In a large study using 536 brain tissue samples, we successfully tested our proposed microscopic system by training a brain cancer classifier on the generated images. The results of this case study show our system's potential to automate diagnostic tasks in pathology and support pathologists in their work. 
We believe that the developed microscopic design could potentially pave the way for the democratization of high-precision diagnosis in resource-constrained settings while maintaining diagnostic quality comparable to traditional high-end equipment.

\subsubsection*{Acknowledgments}
We thank Alexander Bruce for his assistance with slide scanning at the Digital Imaging Facility, Department of Pathology, Brigham and Women’s Hospital, and Kristina C. Grieco, Raquel Arias-Camison, Nina Thakur, Dominique Ballinger, and Alexa Craig for their assistance in collecting the pathology slides. We thank Mariam Kapanadze and Faith McDonald for their administrative support.

\begin{credits}
\subsubsection{\discintname}
This research was partially supported by the Munich Center of Machine Learning, BaCaTeC, and the EU BigPicture project. K.-H.Y. is supported in part by the National Institute of General Medical Sciences grant R35GM142879, the National Heart, Lung, and Blood Institute grant R01HL174679, the Department of Defense Peer Reviewed Cancer Research Program Career Development Award HT9425-231-0523, the Research Scholar Grant RSG-24-1253761-01-ESED (grant DOI: \url{https://doi.org/10.53354/ACS.RSG-24-1253761-01-ESED.pc.gr.193749}) from the American Cancer Society, and the Harvard Medical School Dean's Innovation Award.
\end{credits}
\bibliographystyle{ieeetr}
\bibliography{refs}
\end{document}